\documentclass[11pt]{article}

\usepackage[final]{acl}

\usepackage{times}
\usepackage{latexsym}
\usepackage{amsmath}
\usepackage{amssymb}

\usepackage[T1]{fontenc}

\usepackage[utf8]{inputenc}

\usepackage{microtype}

\usepackage{inconsolata}

\usepackage{graphicx}
\usepackage{url}

\title{Beyond Initialization Loss: A Systematic Study of Token Embedding Initialization Strategies for LLM Vocabulary Extension}

\author{
Raviraj Joshi\thanks{Corresponding author.}, Utkarsh Vaidya, Sanjay Singh Chauhan, Niranjan Wartikar \\
NVIDIA \\
\texttt{\{ravirajj, uvaidya, schauhan, nwartikar\}@nvidia.com}
}

\begin{document}
\maketitle
\begin{abstract}

Vocabulary extension is an efficient way to adapt pretrained large language models (LLMs) to new languages, but the initialization of newly added token embeddings can strongly affect continued pre-training (CPT) efficiency. We present a systematic study of more than $20$ initialization strategies for Hindi vocabulary extension in Nemotron-3-Nano-30B-A3B. Our comparison spans vocabulary-averaging baselines; external and learned initialization methods, including FOCUS, top-$k$ semantic retrieval, and residual MLP mappings; subword composition; norm calibration; and input-output asymmetry. We find that subword composition methods outperform both vocabulary averaging and external/learned initialization approaches. Within subword composition, asymmetric variants achieve the lowest observed early validation loss and reveal distinct preferences for input and output embedding initialization. The best observed configuration initializes $\mathbf{E}_{\text{in}}$ with uniform subword averaging and Hindi-specific norm calibration, and $\mathbf{E}_{\text{out}}$ with character-length-weighted subword averaging. Relative to the standard Mean-all baseline, this full initialization pipeline reaches comparable validation loss with over a $6\times$ reduction in CPT steps and exceeds the baseline's $3{,}500$-step MILU-Hindi accuracy after only $500$ steps. Finally, we show that initialization loss and initialization bits-per-byte (Init BPB) are unreliable predictors of downstream convergence, whereas lightweight CPT---as few as $50$ steps---provides a cost-effective and reliable signal for selecting the best initialization strategy.
\end{abstract}

\section{Introduction}

Modern large language models (LLMs) process text using subword tokenizers such as Byte-Pair Encoding (BPE) \cite{sennrich2015neural} or SentencePiece \cite{kudo-richardson-2018-sentencepiece}, which segment raw text into discrete vocabulary indices. To maximize compression within a fixed vocabulary budget $|V|$, these tokenizers are optimized on large pre-training datasets. Because such datasets are often English-dominant, the resulting vocabularies allocate limited capacity to low-resource languages and non-Latin scripts. When processing these underrepresented languages, the tokenizer frequently fragments words into character- or byte-level pieces. Tokenization efficiency is commonly quantified by fertility, the average number of tokens produced per word, where lower is better. While English typically has a low fertility rate close to $1.0$ token per word, morphologically rich languages, particularly Indic scripts such as Devanagari, are often severely over-tokenized; in these cases, many models exhibit fertility rates exceeding $2.0$ tokens per word \cite{singh-etal-2024-indicgenbench,rust2021good}.

This tokenization asymmetry imposes a substantial tax on low-resource and non-Latin scripts. First, high token fertility inflates the context length $L$ for equivalent semantic content and increases both linear and quadratic pre-training and inference costs. Second, it exhausts the model's fixed context window (e.g., $8192$ tokens), shortening the effective context span available for the target language. Third, it creates an economic burden when API pricing is based on token count: non-Western scripts can incur up to a $5\times$ higher price for comparable semantic content \cite{ahia-etal-2023-languages}. Vocabulary extension addresses this issue by expanding the tokenizer's native vocabulary with language-specific subwords and frequent terms extracted from target-language corpora, thereby reducing sequence lengths and token fertility. In the original Nemotron-3-Nano-30B-A3B tokenizer, only $1{,}569$ of $131{,}072$ vocabulary entries contain Devanagari characters ($1.2\%$), underscoring the limited native script coverage before extension. In our setting, we add $25{,}600$ Hindi tokens sourced directly from the Nanda tokenizer \cite{choudhury2025llama}, which extends the Llama-3 tokenizer with Hindi tokens.
Measured on $100{,}000$ Sangraha Hindi documents~\cite{khan-etal-2024-indicllmsuite}, this extension reduces Hindi token fertility from $1.95$ to $1.25$ tokens per word.

However, while vocabulary extension mitigates the fertility tax, it introduces a parameter optimization challenge at the start of continued pre-training (CPT). Expanding the tokenizer requires resizing the model's embedding matrices, specifically the input lookup matrix ($\mathbf{E}_{\text{in}}$) and the output language modeling head projection ($\mathbf{E}_{\text{out}}$). This structural modification introduces newly allocated vectors that must align with the pretrained embedding manifold and function correctly within the network. Because these parameters have never been updated, they initially lack coherent representations. Poor initialization, including random Gaussian initialization or mean-based Gaussian initialization \cite{hewitt-2021-initializing}, can create a ``cold-start'' bottleneck in which early CPT spends substantial compute aligning new vectors rather than improving target-language modeling. Suboptimal initialization of $\mathbf{E}_{\text{out}}$ can also distort the output distribution, producing unstable logits and high initial loss.

To systematically study this parameter space, we evaluate more than $20$ initialization configurations across five design axes: composition scope, subword weighting, norm calibration, initialization asymmetry, and external encoder projection. Using a Hindi vocabulary extension of Nemotron-3-Nano-30B-A3B, we compare these strategies under both static initialization metrics (initialization loss, or step-zero loss) and short-horizon CPT. Our results support a simple initialization strategy selection protocol in which a lightweight CPT probe of as few as $50$ steps (approximately $0.42$B tokens) is run before committing to a full training run.

We consider vocabulary-averaging baselines, external and learned mapping systems such as FOCUS \cite{dobler-de-melo-2023-focus}, top-$k$ semantic retrieval \cite{minixhofer-etal-2022-wechsel}, residual MLP transformations, and several subword composition and norm calibration schemes. Among these, subword composition is the strongest family of methods. It initializes each new token as a weighted average of the embeddings of its constituent subwords under a chosen weighting heuristic. Building on this observation, we study asymmetric initialization: $\mathbf{E}_{\text{in}}$ and $\mathbf{E}_{\text{out}}$ are initialized with different composition rules because lookup and generation impose different constraints~\cite{adaptivocab-2025}. The best observed configuration initializes $\mathbf{E}_{\text{in}}$ with uniform subword averaging plus language-specific norm calibration, and initializes $\mathbf{E}_{\text{out}}$ with character-length-weighted subword averaging. MuRIL-BERT semantic input weighting paired with character-length output weighting performs nearly identically, suggesting that auxiliary semantic embeddings are also useful for initialization.

Our core contributions are summarized as follows:

\begin{itemize}
    \item \textbf{Comprehensive empirical survey:} We benchmark more than $20$ configurations spanning vocabulary-averaging baselines, external cross-lingual encoders, targeted norm calibration, and asymmetric input-output mappings.
    \item \textbf{Decomposed initialization gains:} We find that subword composition provides the largest improvement over vocabulary-averaging baselines, while input-side Hindi norm calibration and input-output decoupling further refine performance within the stronger subword-composition family. The best observed configuration uses uniform weighting with Hindi norm calibration for $\mathbf{E}_{\text{in}}$ and character-length weighting for $\mathbf{E}_{\text{out}}$, and a near-tie from semantic BERT-based input weighting shows that auxiliary semantic embeddings are useful input-side signals.
    \item \textbf{Lightweight CPT initialization-strategy selection:} We show that initialization-time rankings can change after a brief phase of CPT. A $50$-step probe provides a cost-effective and reliable signal for selecting the best initialization strategy.
    \item \textbf{Downstream acceleration:} Relative to the standard default baseline, the full initialization pipeline of the best observed configuration reaches comparable validation loss with over a $6\times$ CPT-step reduction and exceeds the baseline's $3{,}500$-step MILU-Hindi accuracy after $500$ steps, with a $+25.01$ point absolute gain at the first downstream checkpoint.
\end{itemize}

\section{Related Work}
 
\paragraph{Tokenization Inequity and Vocabulary Extension.}
Modern LLMs rely on subword tokenization algorithms such as Byte-Pair Encoding (BPE)~\cite{sennrich2015neural} and SentencePiece~\cite{kudo-richardson-2018-sentencepiece}, which are trained on corpus statistics and inherently favor high-resource, Latin-script languages. This structural bias manifests as elevated token fertility for morphologically rich or non-Latin scripts. \citet{ahia-etal-2023-languages} quantify the downstream economic consequence, demonstrating across 22 typologically diverse languages that per-token API pricing systematically overcharges speakers of high-fertility languages by up to $5\times$ relative to English. \citet{petrov-etal-2023-language} further formalize this disparity as a ``language tax'' baked into the tokenizer itself. Vocabulary extension~\cite{liu-etal-2023-efficient, gee-etal-2022-fast} directly addresses this problem by augmenting an existing tokenizer with language-specific subwords, reducing fertility and thus lowering both computational and economic costs. Several language-specific LLM efforts have followed this pipeline for Chinese~\cite{cui-etal-2023-efficient}, Japanese~\cite{fujii-etal-2024-continual}, Finnish~\citep{luukkonen-etal-2023-fingpt}, and Korean~\citep{vo-etal-2024-redwhale}.
 
\paragraph{Continued Pre-training for Language Adaptation.}
Vocabulary extension necessitates a phase of continued pre-training (CPT) to integrate the newly added tokens into the pretrained model. A growing body of work has applied this approach to adapt English-centric LLMs to Indic and other low-resource languages. \citet{joshi-etal-2024-nemotron} extend Nemotron-Mini-4B with CPT on 400 billion Hindi-English tokens, reporting state-of-the-art performance on IndicXTREME and related benchmarks. \citet{khan-etal-2024-indicllmsuite} provide a comprehensive blueprint for building pre-training and fine-tuning datasets for 22 Indian languages, and similar multilingual adaptation efforts have been conducted for Turkish~\citep{toraman-2024-llamaturk}, Finnish~\citep{luukkonen-etal-2023-fingpt}, and Korean~\citep{vo-etal-2024-redwhale}. A unifying observation across these works is that initialization quality determines how efficiently the model can absorb new token representations during CPT; nevertheless, most studies treat it as a secondary design choice and default to simple averaging.
 
\paragraph{Embedding Initialization for New Tokens.}
Initializing embeddings for newly added tokens is a well-studied problem in cross-lingual model transfer. The simplest and most widely adopted baseline, proposed by \citet{hewitt-2021-initializing} and integrated as the default in the Hugging Face Transformers library, draws new embeddings from a Gaussian distribution centered at the mean of all existing token embeddings. \citet{gee-etal-2022-fast} and \citet{liu-etal-2023-efficient} show that initializing new tokens as the mean of their constituent subword embeddings accelerates convergence over random initialization. \citet{minixhofer-etal-2022-wechsel} propose WECHSEL, which leverages multilingual static word embeddings and bilingual dictionaries to semantically align new subword embeddings with a source model. Building on this, \citet{dobler-de-melo-2023-focus} propose FOCUS, which represents each new token as a sparse convex combination of overlapping source vocabulary entries via cosine similarity in an auxiliary fastText space and the Sparsemax operator~\cite{martins-astudillo-2016-sparsemax}, removing the need for bilingual dictionaries. More recently, hypernetwork-based approaches~\cite{minixhofer-etal-2024-zero} and embedding factorization methods~\cite{liu-etal-2024-ofa} have been proposed for zero-shot tokenizer transfer and multilingual alignment. The AdaptiVocab work~\cite{adaptivocab-2025} notes that uniform mean initialization is suboptimal for autoregressive generation because it does not account for the structural role a token plays in the output distribution, and therefore applies distinct positional weighting schemes to input embeddings and the LM head. \citet{yamaguchi-etal-2024-empirical} similarly find that auxiliary-encoder-based methods can underperform mean initialization in low-resource regimes, highlighting the distinct sensitivity of the LM head $\mathbf{E}_{\text{out}}$ to its initialization.
 
\paragraph{Evaluation of Initialization Quality.}
A recurring challenge in this line of work is choosing reliable evaluation metrics at initialization time. Initialization-time perplexity and bits-per-byte are commonly used as proxies for downstream performance~\cite{dobler-de-melo-2023-focus, minixhofer-etal-2022-wechsel}, but several studies have noted that rankings over these static metrics can be misleading. \citet{gee-etal-2022-fast} observe that initialization quality does not always translate directly to downstream fine-tuning performance, and \citet{yamaguchi-etal-2024-empirical} find that the relative ordering of strategies can shift after even a short training phase. This motivates evaluation protocols that go beyond step-zero metrics.
 
\section{Baselines and External Encoder Frameworks}
We evaluate two standard vocabulary-averaging methods alongside several external mapping paradigms. Unless otherwise stated, each method is applied symmetrically to both $\mathbf{E}_{\text{in}}$ and $\mathbf{E}_{\text{out}}$.

\paragraph{Mean-All (Default Baseline):} Following tokenizer vocabulary expansion, newly added token embeddings are initialized using the mean embedding initialization method introduced by \citet{hewitt-2021-initializing} and adopted as the default in the Hugging Face Transformers library (\texttt{mean\_resizing=True}). Concretely, given the pre-expansion embedding matrix $\mathbf{E} \in \mathbb{R}^{n \times d}$, the mean embedding $\boldsymbol{\mu}$ and empirical covariance $\boldsymbol{\Sigma}$ are computed as:
\[
\boldsymbol{\mu} = \frac{1}{n} \sum_{i=1}^{n} \mathbf{e}_i, \qquad \boldsymbol{\Sigma} = \frac{(\mathbf{E} - \boldsymbol{\mu})^\top (\mathbf{E} - \boldsymbol{\mu})}{n}
\]
where $\boldsymbol{\mu}$ is broadcast across rows. Each new token embedding is then sampled from a scaled multivariate normal distribution:
\[
\mathbf{e}_{\text{new}} \sim \mathcal{N}\left(\boldsymbol{\mu},\; \alpha \boldsymbol{\Sigma}\right)
\]
where $\alpha$ is a small scaling factor (e.g., $\alpha = 10^{-5}$).

\paragraph{Mean-Hindi (Script-Restricted Baseline):} Structurally identical to the Mean-All baseline, this method computes the localized mean $\boldsymbol{\mu}_{\text{Hi}}$ and empirical covariance $\boldsymbol{\Sigma}_{\text{Hi}}$ by restricting the source token pool strictly to pre-existing vocabulary items that contain Devanagari script characters ($V_{\text{Hi}} \subset V$). Devanagari script tokens are identified dynamically using a regular expression matching filter that scans the decoded token strings for characters within the Devanagari Unicode range $[\text{U+0900}, \text{U+097F}]$. New embeddings are then sampled using the covariance-scaled multivariate normal distribution:
\[
\mathbf{e}_{\text{new}} \sim \mathcal{N}\left(\boldsymbol{\mu}_{\text{Hi}},\; \alpha \boldsymbol{\Sigma}_{\text{Hi}}\right)
\]

\paragraph{FOCUS (Sparsemax-weighted fastText Alignment):} We implement FOCUS~\cite{dobler-de-melo-2023-focus} to initialize new tokens as localized, sparse convex combinations of base token embeddings. Specifically, for each new token $\tau$, we extract semantic vectors $\mathbf{f}(\tau) \in \mathbb{R}^{300}$ from a Hindi fastText model~\cite{bojanowski2017enriching} (\texttt{cc.hi.300.bin}). This deviates from the original FOCUS recipe, which trains fastText on target-language data tokenized with the new tokenizer; we use public Hindi fastText vectors for reproducibility and lower auxiliary-training cost. We compute its cosine similarity against all baseline candidates $c$ in a candidate pool $\mathcal{C}$ and scale the logits using a sharpening temperature $\tau_{\text{temp}} = 0.05$ to focus probability mass:
\[
z_c = \frac{\mathbf{f}(\tau) \cdot \mathbf{f}(c)}{\tau_{\text{temp}} \|\mathbf{f}(\tau)\|_2 \|\mathbf{f}(c)\|_2}
\]
These scaled logits $\mathbf{z} \in \mathbb{R}^{|\mathcal{C}|}$ are mapped to a sparse probability distribution $\mathbf{w}$ using the Sparsemax activation function~\cite{martins-astudillo-2016-sparsemax}, which projects the inputs onto the $(|\mathcal{C}|-1)$-dimensional simplex and sets lower-scoring candidates to exactly $0$:
\[
\mathbf{w} = \operatorname{sparsemax}(\mathbf{z}) = \operatorname*{argmin}_{\mathbf{p} \in \Delta^{|\mathcal{C}|-1}} \|\mathbf{p} - \mathbf{z}\|_2^2
\]
We evaluate two distinct candidate pools $\mathcal{C}$: standard FOCUS (\texttt{ft-hi}), where $\mathcal{C} = V$; and Anchor FOCUS (\texttt{ft-hi, anchor-hi}), where the pool is restricted to original Hindi/Devanagari vocabulary items ($\mathcal{C} = V_{\text{Hi}}$). The final new token embeddings are synthesized as:
\[
\mathbf{e}(\tau) = \sum_{c \in \mathcal{C}} w_c \cdot \mathbf{e}(c)
\]

\paragraph{Top-$k$ Gemma Retrieval (Weighted Top-$k$ Alignment):} 
Following the retrieval-and-transfer paradigm of WECHSEL~\cite{minixhofer-etal-2022-wechsel}, this method constructs a representation for each new token $\tau$ by identifying its $k=5$ nearest neighbors in an auxiliary LLM embedding space and transferring a weighted combination of their pretrained embeddings. Rather than fastText vectors, we use the static input embedding layer of Gemma-2-27B~\cite{team2024gemma} as the auxiliary space. Gemma-2-27B is chosen for its stronger multilingual and subword-level representations compared to static word vectors, and for its Indic token coverage.

We first extract a mean-pooled Gemma embedding $\mathbf{g}(\tau) \in \mathbb{R}^{4608}$ across its Gemma-tokenized subwords:
\[
\mathbf{g}(\tau) = \frac{\sum_{t=1}^T M_t \cdot \mathbf{E}_{\text{Gemma}}(x_t)}{\sum_{t=1}^T M_t}
\]
where $M_t \in \{0, 1\}$ is the attention mask. We compute $\mathbf{g}(v)$ analogously for each original vocabulary token $v \in V$, then calculate cosine similarities in the Gemma space to find the top-$k$ nearest neighbors $\mathcal{K}_{\tau} = \{i_1, \dots, i_k\}$. To map similarities to non-negative composition weights, we shift and normalize them relative to their minimum value in the subset:
\begin{align*}
\operatorname{sim}'_j &= \operatorname{sim}(\tau, i_j) - \min_{m \in \mathcal{K}_{\tau}} \operatorname{sim}(\tau, m) + \epsilon \\[0.5em]
w_j &= \frac{\operatorname{sim}'_j}{\sum_{m=1}^k \operatorname{sim}'_m}
\end{align*}
where $\epsilon = 10^{-8}$ is a stability constant. The uncalibrated representation is synthesized as $\mathbf{e}_{\text{avg}}(\tau) = \sum_{j=1}^k w_j \cdot \mathbf{e}(i_j)$. To reconcile magnitude shrinkage on the lookup side without increasing output logit instability, we selectively apply norm calibration to the input embedding space ($\mathbf{E}_{\text{in}}$) while bypassing the output language modeling head ($\mathbf{E}_{\text{out}}$):
\[
\mathbf{e}_{\text{cal, in}}(\tau) = \tilde{\mu}_{\text{in}} \cdot \frac{\mathbf{e}_{\text{avg, in}}(\tau)}{\|\mathbf{e}_{\text{avg, in}}(\tau)\|_2}
\]

\paragraph{Residual MLP with Mean-Centered Alignment (Residual MLP):} We implement a supervised residual mapping designed to reduce the mismatch between retrieval-based proxy vectors and the pretrained embedding space. 

The pipeline is executed through the following sequence:

\textbf{Anchor Preparation:} We define a set of native Hindi anchor tokens $\mathcal{A} \subset V_{\text{Hi}}$ with known pretrained embeddings $\mathbf{y}_a \in \mathbb{R}^d$. For each anchor, we obtain an approximate proxy vector $\mathbf{x}_a \in \mathbb{R}^d$ via top-$k$ similarity retrieval constrained strictly to non-Hindi baseline tokens. To eliminate structural translation shifts between these spaces, we calculate centroids $\boldsymbol{\mu}_X$ and $\boldsymbol{\mu}_Y$ over $\mathcal{A}$ to mean-center the representations:
\[
\mathbf{x}_a^c = \mathbf{x}_a - \boldsymbol{\mu}_X, \qquad \mathbf{y}_a^c = \mathbf{y}_a - \boldsymbol{\mu}_Y
\]

\textbf{Network Architecture and Objective:} A residual mapping model $f(\mathbf{x}_a^c) = \mathbf{x}_a^c + \operatorname{MLP}(\mathbf{x}_a^c)$ is parameterized by an MLP consisting of a Layer Normalization layer, a hidden layer of dimension $d_{\text{hidden}} = 1024$, a GELU activation function, a dropout rate of $0.1$, and a final linear projection back to $d$ dimensions. The network is trained by minimizing a multi-component loss function over the batch size $B$:
\begin{align*}
    \mathcal{L} &= \mathcal{L}_{\text{MSE}} + \lambda_{\text{cos}} \mathcal{L}_{\text{cos}} + \lambda_{\text{norm}} \mathcal{L}_{\text{norm}} \\[0.5em]
    \text{where:} \quad \mathcal{L}_{\text{MSE}} &= \frac{1}{B} \sum_{b=1}^{B} \|f(\mathbf{x}_b^c) - \mathbf{y}_b^c\|_2^2 \\
    \mathcal{L}_{\text{cos}} &= 1 - \frac{1}{B} \sum_{b=1}^{B} \frac{f(\mathbf{x}_b^c) \cdot \mathbf{y}_b^c}{\|f(\mathbf{x}_b^c)\|_2 \|\mathbf{y}_b^c\|_2} \\
    \mathcal{L}_{\text{norm}} &= \frac{1}{B} \sum_{b=1}^{B} \left( \|f(\mathbf{x}_b^c)\|_2 - \|\mathbf{y}_b^c\|_2 \right)^2
\end{align*}
where $\lambda_{\text{cos}} = 0.5$ and $\lambda_{\text{norm}} = 0.1$.

\textbf{Inference Pipeline:} For an unseen token $\tau$ with proxy representation $\mathbf{x}_{\tau}$ (retrieved via non-Hindi top-$k$ baseline tokens), we predict its initialized embedding vector as:
\[
\mathbf{e}(\tau) = f(\mathbf{x}_{\tau} - \boldsymbol{\mu}_X) + \boldsymbol{\mu}_Y
\]

\section{Subword Composition Mechanics}
Prior work has established subword averaging as a simple yet effective initialization strategy for vocabulary extension~\cite{gee-etal-2022-fast, liu-etal-2023-efficient}: a new token's embedding is initialized as the mean of its constituent subword embeddings from the original vocabulary. However, these approaches treat all subwords as equally weighted and apply a symmetric strategy to both $\mathbf{E}_{\text{in}}$ and $\mathbf{E}_{\text{out}}$, leaving the broader design space of subword composition largely unexplored. We generalize this idea along two axes: (i) the weighting scheme applied to constituent subwords, and (ii) the decoupling of initialization strategies between the input embedding matrix and the output language modeling head.

Formally, for any newly introduced token $\tau$ that decomposes into a sequence of $n$ valid subwords $s_1, s_2, \dots, s_n$ within the original vocabulary $V$, we initialize its representation as a weighted linear combination of the subword embeddings:
\[
\mathbf{e}(\tau) = \sum_{i=1}^{n} w_i \cdot \mathbf{e}(s_i)
\]
This weighted formulation can be configured independently for the input embedding matrix $\mathbf{E}_{\text{in}}$ and the output language modeling head matrix $\mathbf{E}_{\text{out}}$ (asymmetric composition). We examine four weighting families, yielding five implemented variants, for deriving the weighting vector $\mathbf{w} \in \mathbb{R}^n$:

\begin{itemize}
    \item \textbf{Uniform:} Assigns equal contribution to all constituent subwords:
    \[
    w_i = \frac{1}{n}
    \]
    
    \item \textbf{Character-Length (Char-len):} Scales weights proportionally to the textual length of the decoded subwords, giving longer fragments greater influence:
    \[
    w_i = \frac{|s_i|}{\sum_{j=1}^{n} |s_j|}
    \]
    where $|s_i| = \max(\operatorname{len}(\operatorname{decode}(s_i)), 1)$ represents the decoded character length of subword $s_i$. In our implementation, $\operatorname{len}(\cdot)$ counts Unicode code points in the tokenizer-decoded string, without additional Unicode normalization.
    
    \item \textbf{Max-Character (Max-char):} A sparse, one-hot selection that copies only the embedding of the longest constituent subword:
    \[
    w_i = \begin{cases} 1 & \text{if } i = \operatorname*{argmax}_{j} |s_j| \\ 0 & \text{otherwise} \end{cases}
    \]
    To handle length equivalence rigorously, ties are broken deterministically by selecting the first occurrence in the subword sequence.
    
    \item \textbf{Semantic Similarity (MuRIL-weighted / Gemma-weighted):} Computes semantic weights by matching the full token $\tau$ with its decomposed fragments in an auxiliary representation space. Let $\mathbf{h}(\tau)$ be the semantic embedding of the full clean token, and $\mathbf{h}(s_i)$ be the semantic embedding of subword $s_i$. The raw alignment score is defined by cosine similarity:
    \[
    \operatorname{sim}_i = \frac{\mathbf{h}(\tau) \cdot \mathbf{h}(s_i)}{\| \mathbf{h}(\tau)\|_2 \| \mathbf{h}(s_i)\|_2}
    \]
    The subword weights are derived by passing the scaled similarity scores through a softmax function parameterized by temperature $\tau_{\text{temp}} > 0$:
    \[
    w_i = \frac{\exp(\operatorname{sim}_i / \tau_{\text{temp}})}{\sum_{j=1}^{n} \exp(\operatorname{sim}_j / \tau_{\text{temp}})}
    \]
    where $\tau_{\text{temp}} = 0.1$ encourages sharp distributions. We investigate two representation extractors for mapping $\mathbf{h}(\cdot)$:
    
    \begin{itemize}
        \item \textbf{MuRIL-based (MuRIL-weighted):} Leverages Google's Multilingual Representations for Indian Languages~\cite{khanuja-etal-2021-muril} (\texttt{google/muril-base-cased}). Rather than static lookups, the representation $\mathbf{h}(\mathbf{x})$ is extracted from the model's final encoder layer (\texttt{last\_hidden\_state}) via attention-mask-aware mean pooling:
        \[
        \mathbf{h}(\mathbf{x}) = \frac{\sum_{t=1}^T M_t \cdot \mathbf{z}_t}{\sum_{t=1}^T M_t}
        \]
        where $\mathbf{z}_t$ represents the final hidden state vector at sequence position $t$, and $M_t \in \{0, 1\}$ represents the attention mask.
        
        \item \textbf{Gemma-based (Gemma-weighted):} Leverages Gemma-2-27B (\texttt{google/gemma-2-27b}). To minimize compute overhead, it maps representations purely using the static input embedding layer (\texttt{embed\_tokens}) pooled uniformly over non-padding tokens:
        \[
        \mathbf{h}(\mathbf{x}) = \frac{1}{T} \sum_{t=1}^T \mathbf{E}_{\text{Gemma}}(x_t)
        \]
        where $\mathbf{E}_{\text{Gemma}}$ denotes the static token embedding lookup.
    \end{itemize}
\end{itemize}

\paragraph{Decoupling the Matrices (Symmetric vs. Asymmetric)}
\begin{itemize}
    \item \textbf{Symmetric Initialization:} Applies an identical weighting paradigm to both the input embedding matrix ($\mathbf{E}_{\text{in}}$) and the output language modeling head ($\mathbf{E}_{\text{out}}$).
    \item \textbf{Asymmetric Initialization:} Decouples these spaces entirely, allowing distinct mathematical formulations for token lookup ($\mathbf{E}_{\text{in}}$) and token prediction ($\mathbf{E}_{\text{out}}$).
\end{itemize}

\subsection{Norm Calibration Dynamics}
Subword averaging tends to reduce the $L_2$ norm of synthesized embeddings relative to native token embeddings, since averaging vectors that do not perfectly align results in a smaller combined magnitude. The resulting norm mismatch can produce lower logits and uneven gradient magnitudes, causing the model to underutilize new tokens early in training and ultimately slowing convergence. To rectify this structural deficit, we apply norm calibration:
\[
\mathbf{e}_{\text{cal}}(\tau) = \tilde{\mu} \cdot \frac{\mathbf{e}(\tau)}{\|\mathbf{e}(\tau)\|_2}
\]
where $\tilde{\mu}$ represents the target calibration norm. We define and evaluate two alternative formulations for $\tilde{\mu}$ based on the median $L_2$ norm of an original vocabulary subset ($V_{\text{subset}}$):
\[
\tilde{\mu} = \operatorname*{median}_{v \in V_{\text{subset}}} \left( \|\mathbf{e}(v)\|_2 \right)
\]

\begin{itemize}
    \item \textbf{Global Norm Calibration:} Rescales new vectors to match the median norm calculated across the entire baseline vocabulary ($V_{\text{subset}} = V$).
    \item \textbf{Language-Specific Hindi Norm Calibration:} Restricts the target median calculation strictly to active Hindi/Devanagari tokens within the original vocabulary ($V_{\text{subset}} = V_{\text{Hi}}$). Script membership is identified by scanning for Unicode characters falling within the Devanagari range (\texttt{U+0900} to \texttt{U+097F}).
\end{itemize}
In norm-calibrated configurations that we carry forward, calibration is applied only to the input matrix $\mathbf{E}_{\text{in}}$. We include an input-output calibration ablation, but rescaling $\mathbf{E}_{\text{out}}$ sharply increases initialization loss, consistent with output logit inflation. Although this variant partially recovers after short CPT, it is not competitive with input-only calibration.

\subsection{Evaluation Protocol}
Prior literature often relies on initialization-time perplexity or cross-entropy loss to evaluate initialization strategies. However, we find that initialization loss and initialization bits-per-byte (\textit{Init BPB}) are not sufficient for choosing the best configuration. Performance rankings frequently shift during the initial phase of CPT. We therefore use a lightweight CPT run of $50$ steps for initialization-strategy selection. Empirical results indicate that relative rankings stabilize after these $50$ steps, making validation loss at this stage ($\text{Val}_{50}$) a cost-effective signal for selecting the initialization strategy, with training-loss trajectories showing similar early trends.

\section{Empirical Evaluation and Discussion}
All experiments are conducted using the Nemotron-3-Nano-30B-A3B architecture, with CPT data drawn from the Hindi subset of the Nemotron v3 pretraining blend\footnote{\url{https://huggingface.co/collections/nvidia/nemotron-v3-pre-training}} and evaluation metrics reported on a held-out Hindi evaluation corpus.

\paragraph{Experimental setup.}
All short CPT probes use the same $1:1$ blend of Hindi CPT data and the original MultiMix English/code/math/STEM data as the longer runs: global batch size $1024$, sequence length $8192$, AdamW-style optimization with peak learning rate $1\times10^{-5}$, cosine decay, and $128$ NVIDIA A100 GPUs across $16$ nodes. The model uses untied input and output embeddings, enabling asymmetric initialization of $\mathbf{E}_{\text{in}}$ and $\mathbf{E}_{\text{out}}$. After initialization, all parameters are trained jointly without a PEFT stage; a $50$-step probe processes approximately $0.42$B tokens, which is non-trivial but less than $1\%$ of the $7{,}000$-step CPT run at the same batch size.

\begin{table*}[t]
\centering
\scriptsize
\setlength{\tabcolsep}{3pt}
\caption{Empirical profiles across baselines, external encoders, symmetric subword/norm ablations, and asymmetric initialization configurations. Init Loss denotes cross-entropy measured immediately after embedding initialization and before CPT; Init BPB denotes bits-per-byte at step $0$; $\text{Val}_{50}$ denotes validation loss after $50$ CPT steps. Bold values indicate the best result within each block and metric column.}
\label{tab:consolidated_evals}
\resizebox{\textwidth}{!}{
\begin{tabular}{lcccccc}
\noalign{\hrule height 1.3pt}
\textbf{Method} & \textbf{Input Wt.} & \textbf{Output Wt.} & \textbf{Norm} & \textbf{Init Loss} & \textbf{Init BPB} & \textbf{Val$_{50}$} \\
\noalign{\hrule height 1.3pt}
\multicolumn{7}{l}{\textbf{A. Baselines and External Encoders}} \\ \hline
Mean-all & --- & --- & --- & 13.346 & 1.8706 & 5.527 \\ \hline
Mean-Hindi & --- & --- & --- & 10.878 & 1.5247 & 4.736 \\ \hline
FOCUS (ft-hi) & --- & --- & --- & 8.080 & 1.1325 & --- \\ \hline
FOCUS (anchor-hi) & --- & --- & --- & \textbf{7.549} & \textbf{1.0581} & 3.276 \\ \hline
Top-$k$ Gemma & --- & --- & --- & 8.098 & 1.1351 & \textbf{3.222} \\ \hline
Residual MLP & --- & --- & --- & 12.199 & 1.7098 & 3.656 \\ \hline

\multicolumn{7}{l}{\textbf{B. Symmetric Subword Composition and Norm Calibration}} \\ \hline
Uniform & Uniform & Uniform & None & 7.499 & 1.0511 & 2.806 \\ \hline
Char-len & Char-len & Char-len & None & \textbf{7.216} & \textbf{1.0113} & 2.818 \\ \hline
MuRIL-weighted & MuRIL & MuRIL & None & 7.377 & 1.0339 & 2.807 \\ \hline
Gemma-weighted & Gemma & Gemma & None & 7.725 & 1.0827 & 3.017 \\ \hline
Max-char & Max-char & Max-char & None & 8.217 & 1.1517 & 3.284 \\ \hline
Global Norm & Uniform & Uniform & Global (in) & 7.379 & 1.0343 & 2.758 \\ \hline
Hindi Norm & Uniform & Uniform & Hindi (in) & 7.372 & 1.0332 & \textbf{2.753} \\ \hline
Hindi Norm (in+out) & Uniform & Uniform & Hindi (in+out) & 17.618 & 2.4694 & 2.914 \\ \hline

\multicolumn{7}{l}{\textbf{C. Asymmetric Initialization}} \\ \hline
Symmetric baseline & Uniform & Uniform & Hindi (in) & 7.372 & 1.0332 & 2.753 \\ \hline
Char-len In & Char-len & Uniform & Hindi (in) & 7.266 & 1.0184 & 2.834 \\
MuRIL In & MuRIL & Uniform & Hindi (in) & 7.372 & 1.0332 & 2.756 \\
Gemma In & Gemma & Uniform & Hindi (in) & 7.342 & 1.0290 & 2.850 \\
Max-char In & Max-char & Uniform & Hindi (in) & \textbf{7.246} & \textbf{1.0156} & 3.027 \\ \hline
Char-len Out & Uniform & Char-len & Hindi (in) & 7.256 & 1.0170 & \textbf{2.722} \\
MuRIL Out & Uniform & MuRIL & Hindi (in) & 7.364 & 1.0321 & 2.752 \\
Gemma Out & Uniform & Gemma & Hindi (in) & 7.728 & 1.0831 & 2.927 \\
Max-char Out & Uniform & Max-char & Hindi (in) & 8.160 & 1.1437 & 3.122 \\ \hline
MuRIL In + Char-len Out & MuRIL & Char-len & Hindi (in) & 7.258 & 1.0173 & 2.724 \\
\noalign{\hrule height 1.3pt}
\end{tabular}
}
\end{table*}

\begin{figure}[t]
\centering
\includegraphics[width=\columnwidth]{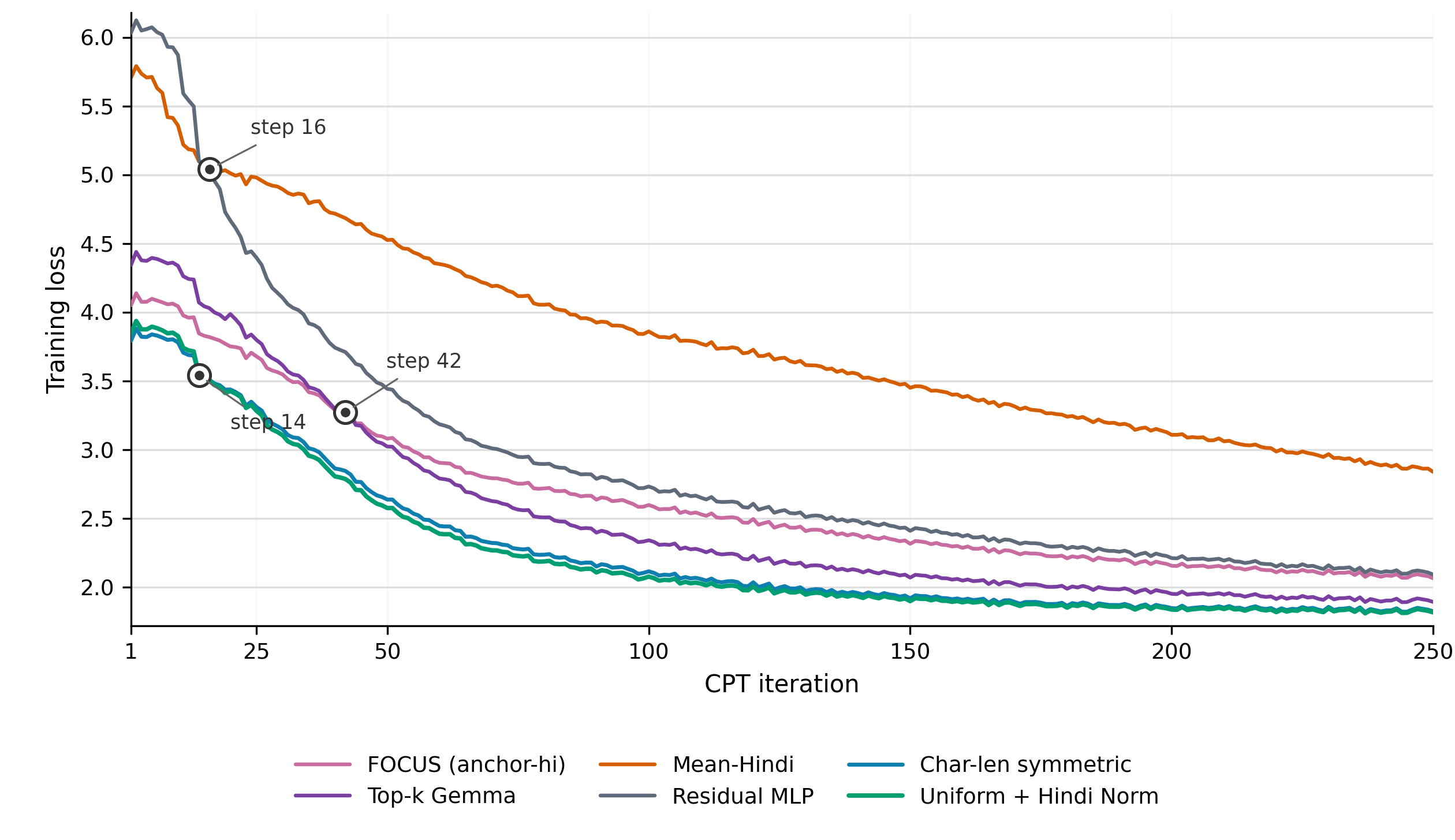}
\caption{Training loss trajectories for representative configurations from Sections A and B whose relative ordering changes during early CPT. Methods favored by static initialization metrics or early training losses, such as \texttt{FOCUS (anchor-hi)}, \texttt{Mean-Hindi}, and symmetric \texttt{Char-len}, are overtaken by alternatives with better early optimization trajectories.}
\label{fig:loss_crossovers}
\end{figure}

\subsection{Baselines and External Encoder Performance}
We first evaluate two vocabulary initialization paradigms: statistical token-averaging baselines (\texttt{Mean-all} and \texttt{Mean-Hindi}) and external cross-lingual mapping schemes, including \texttt{FOCUS} with and without script anchors, \texttt{Top-$k$ Gemma} retrieval, and a residual \texttt{MLP} framework. The initialization cross-entropy loss, initial bits-per-byte (BPB), and early validation trajectory ($\text{Val}_{50}$) for these configurations are reported in Table~\ref{tab:consolidated_evals} (Section A).

Within this suite, anchor-augmented \texttt{FOCUS} yields the lowest initialization error ($\text{Init Loss} = \mathbf{7.549}$). However, the ordering changes after lightweight CPT. At step $0$, \texttt{FOCUS (anchor-hi)} outperforms \texttt{Top-$k$ Gemma} in initialization loss (7.549 vs.\ 8.098). By step $50$, \texttt{Top-$k$ Gemma} achieves the best Section A validation loss ($\text{Val}_{50} = \mathbf{3.222}$).
Figure~\ref{fig:loss_crossovers} shows the corresponding training-loss view for representative methods, where the same qualitative ordering changes appear during early CPT.

The residual \texttt{MLP} shows a similar pattern: it initializes poorly with an $\text{Init Loss}$ of 12.199, worse than the script-restricted \texttt{Mean-Hindi} baseline (10.878), yet reaches a lower $\text{Val}_{50}$ than \texttt{Mean-Hindi} after $50$ steps ($3.656$ vs.\ $4.736$). These inversions indicate that static cross-entropy does not fully determine how quickly an initialization can be optimized. Although restricting the averaging pool to target-script tokens (\texttt{Mean-Hindi}) improves substantially over \texttt{Mean-all}, it remains weaker than the best external mapping methods after short CPT.

\subsection{Symmetric Subword Averaging and Norm Calibration Ablations}
As a computationally efficient alternative to external encoders, we examine two groups of internal methods: symmetric subword composition strategies (\texttt{Uniform}, \texttt{Char-len}, \texttt{MuRIL-weighted}, \texttt{Gemma-weighted}, and \texttt{Max-char}) and norm calibration variants (input-only \texttt{Global Norm}, input-only \texttt{Hindi Norm}, and dual input-output \texttt{Hindi Norm}). The empirical profiles of these configurations are detailed in Table~\ref{tab:consolidated_evals} (Section B).

A clear ranking change emerges between \texttt{Char-len} and \texttt{Uniform + Hindi Norm (Input matrix only)}. The symmetric \texttt{Char-len} heuristic gives the best Section B initialization-time metrics ($\text{Init Loss} = \mathbf{7.216}$; $\text{Init BPB} = \mathbf{1.0113}$). After $50$ steps, however, the \texttt{Uniform + Hindi Norm (Input matrix only)} variant gives the best Section B early optimization result ($\text{Val}_{50} = \mathbf{2.753}$). Input-only norm calibration also outperforms all external baselines from Section A. These results suggest that norm calibration is an important companion to subword composition, especially on the input side where averaged vectors otherwise suffer from reduced embedding norms. Figure~\ref{fig:loss_crossovers} visualizes analogous training-loss curve crossings for representative Section A and Section B configurations during early CPT.

Conversely, extending this calibration symmetrically to the output layer sharply worsens initialization ($\text{Init Loss} = 17.618$). This variant partially recovers under $50$ CPT steps ($\text{Val}_{50} = 2.914$), but remains worse than input-only norm calibration. This functional divergence indicates that input lookup and output prediction benefit from different initialization treatments, motivating the asymmetric framework evaluated next.

\subsection{Asymmetric Initialization Strategies}
We next evaluate decoupled, asymmetric configurations with input-side norm calibration. The experimental matrix includes input-side ablations (varying input strategies while fixing output to \texttt{Uniform}), output-side ablations (varying output strategies while fixing input to \texttt{Uniform}), and a hybrid \texttt{Cross-Strategy Combo} (\texttt{MuRIL} input with \texttt{Char-len} output), as cataloged in Table~\ref{tab:consolidated_evals} (Section C).

The decoupling of layers again reveals a divergence between initialization quality and early trainability. On the input side, the sparse \texttt{Max-char In} mapping achieves the best initialization footprint within the input-side ablation ($\text{Init Loss} = \mathbf{7.246}$; $\text{Init BPB} = \mathbf{1.0156}$). Despite this favorable cold-start state, it reaches only $\text{Val}_{50} = 3.027$, the weakest result within the asymmetric cohort.

Conversely, the output-side ablation pairing a \texttt{Uniform} input strategy with a \texttt{Char-len} output strategy begins with a slightly higher initialization loss ($\text{Init Loss} = 7.256$; $\text{Init BPB} = 1.0170$), but gives the lowest observed overall early validation loss ($\text{Val}_{50} = \mathbf{2.722}$), narrowly ahead of several strong alternatives. These outcomes suggest that input lookup benefits from uniform composition, whereas output prediction benefits from length-proportional weighting. The \texttt{Cross-Strategy Combo} (\texttt{MuRIL-BERT In + Char-len Out}) is nearly tied with the best observed configuration ($\text{Val}_{50} = 2.724$), showing that BERT-style semantic embeddings can also provide useful input-side initialization signals. Selecting a strategy based solely on initialization loss would therefore choose a weaker initializer. Consequently, the best observed configuration, \texttt{Uniform In + Char-len Out + Hindi Input Norm Calibration}, is used for downstream evaluation and large-scale validation, with \texttt{MuRIL-BERT In + Char-len Out} as a close runner-up.

\subsection{Training Convergence Dynamics}
To quantify end-to-end acceleration relative to the default initialization, we compare the best observed asymmetric configuration against the standard \texttt{Mean-all} baseline. At step $50$, this configuration reduces validation cross-entropy by \textbf{$50.8\%$} relative to the baseline ($\mathbf{2.72}$ vs.\ $5.53$). In the extended convergence traces in Figure~\ref{fig:extended_loss_milu}, the baseline reaches a comparable validation loss after approximately $321$ iterations, corresponding to over a \textbf{$6\times$ reduction} in CPT steps for comparable validation loss.
This speedup reflects the full initialization pipeline relative to the default baseline; the component ablations in Table~\ref{tab:consolidated_evals} separately show the contribution of norm calibration and input-output decoupling within the stronger subword-composition family.

\begin{figure}[t]
\centering
\includegraphics[width=\columnwidth]{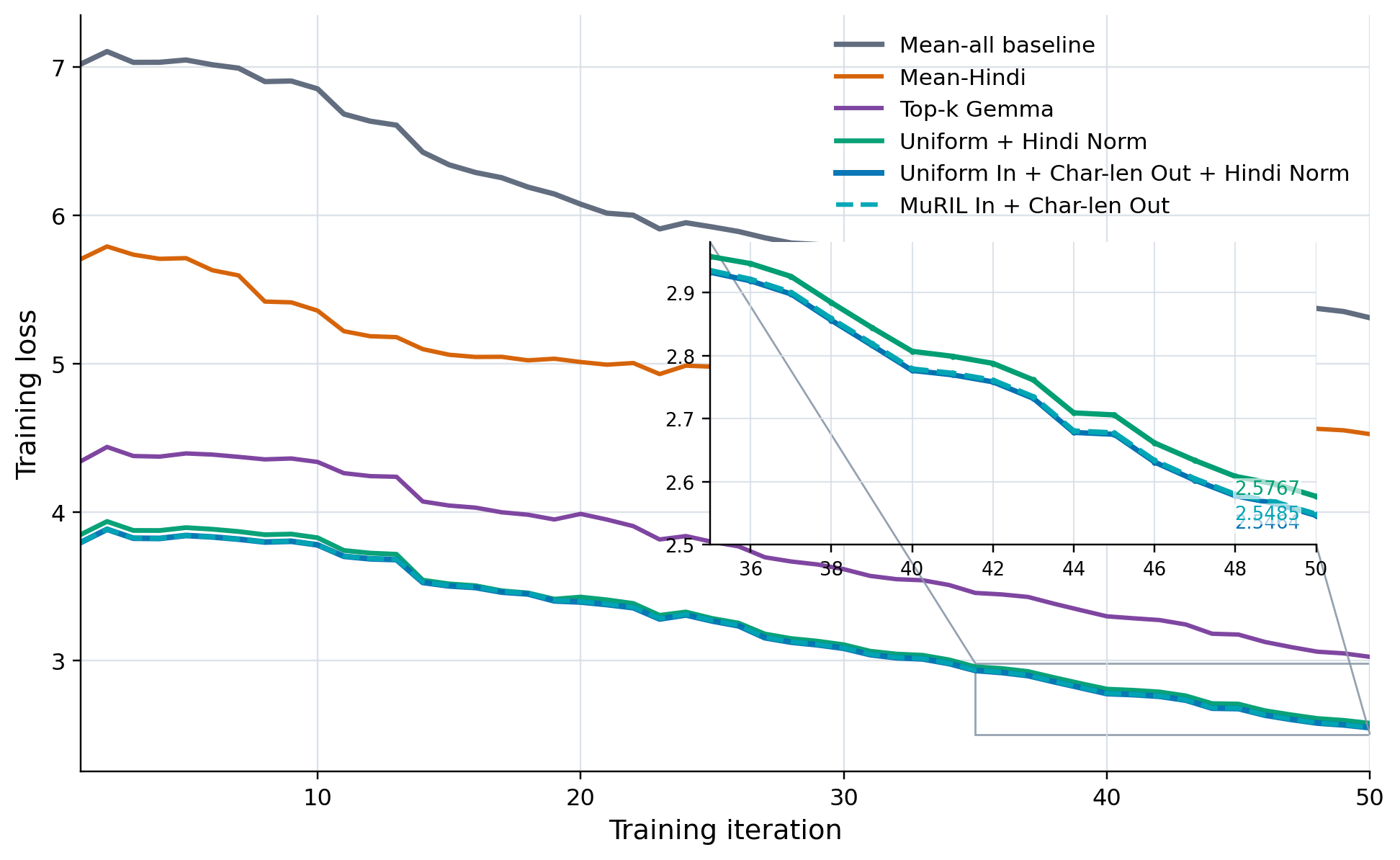}
\caption{Training loss over the first $50$ CPT steps for key configurations. The best observed asymmetric initialization and the \texttt{MuRIL-BERT In + Char-len Out} semantic-input runner-up closely track one another and remain below vocabulary-averaging and external-retrieval baselines during the lightweight CPT probe.}
\label{fig:key_loss_first50}
\end{figure}

The ranking changes observed between $\text{Init Loss}$ and $\text{Val}_{50}$ largely subside beyond this point. In our runs, relative rankings stabilize by step $50$, supporting lightweight continued pre-training as a practical proxy for longer-horizon optimization and as a stable basis for downstream task selection. Figure~\ref{fig:key_loss_first50} shows the corresponding early trajectories for the best observed configuration, its runner-up, and representative baselines.

\subsection{Downstream Benchmark Evaluation (MILU-Hindi)}
To verify that validation cross-entropy improvements translate to functional task enhancements, we evaluate target-language performance against the baseline on MILU-Hindi, the Hindi subset of the Multi-task Indic Language Understanding benchmark~\cite{verma2025milu}, across progressive CPT checkpoints from $500$ to $3,500$ steps (Table~\ref{tab:downstream_milu}).

\begin{table}[h]
\centering
\small
\begin{tabular}{cccc}
\noalign{\hrule height 1.3pt}
\textbf{CPT Step} & \textbf{Baseline} & \textbf{Best Method} & \textbf{Gain ($\Delta$)} \\
\noalign{\hrule height 1.3pt}
500 & 32.00 & \textbf{57.01} & \textbf{+25.01} \\ \hline
1500 & 50.18 & \textbf{60.71} & \textbf{+10.53} \\ \hline
2500 & 54.53 & \textbf{61.80} & \textbf{+7.27} \\ \hline
3000 & 55.80 & \textbf{62.30} & \textbf{+6.50} \\ \hline
3500 & 56.37 & \textbf{62.81} & \textbf{+6.44} \\
\noalign{\hrule height 1.3pt}
\end{tabular}
\caption{Downstream evaluation performance on the MILU-Hindi benchmark across matched CPT checkpoints. Baseline denotes Mean-all; Best Method denotes Uniform In + Char-len Out with Hindi input norm calibration.}
\label{tab:downstream_milu}
\end{table}

The proposed asymmetric framework (\texttt{Best Method}) outperforms the baseline at every checkpoint, reaching \textbf{62.81}. At step $500$, it shows an absolute gain of \textbf{$+25.01$ points} ($+78.2\%$ relative) over the baseline. It also exceeds the baseline's $3{,}500$-step MILU-Hindi performance within $500$ steps, corresponding to a \textbf{$7\times$ reduction} in required training steps for this downstream target.

\subsection{Asymptotic Stability and Scalability: Roughly 50B Continued Pre-training Regime}
In the final experimental phase, we execute an extended continued pre-training run encompassing roughly $50$ billion tokens ($7{,}000$ global steps, $1:1$ Hindi-English ratio) to evaluate long-term stability and source-language retention. We compare the unextended reference model (\texttt{Original}) against the vocabulary-extended model initialized with our best observed method (\texttt{Extended Best}) in Table~\ref{tab:long_run}.

\begin{table}[h]
\centering
\small
\begin{tabular}{lcc}
\noalign{\hrule height 1.3pt}
\textbf{Benchmark Namespace} & \textbf{Original} & \textbf{Extended Best} \\
\noalign{\hrule height 1.3pt}
\textbf{MILU-Hindi} (Target) & 63.56 & \textbf{63.77} (\textbf{+0.21}) \\ \hline
\textbf{MILU-English} (Source) & \textbf{74.79} & 74.38 ($-0.41$) \\
\noalign{\hrule height 1.3pt}
\end{tabular}
\caption{Long-run MILU evaluation. Original denotes the unextended reference model; Extended Best denotes the vocabulary-extended model initialized with the best observed asymmetric method after roughly $50$B tokens of continued pre-training.}
\label{tab:long_run}
\end{table}

The extended model slightly improves MILU-Hindi performance over the original reference model (\textbf{63.77} vs.\ 63.56), while MILU-English remains close to the original (74.38 vs.\ \textbf{74.79}). Appendix~\ref{app:retention_results} provides a broader retention evaluation showing that Extended Best matches the original within evaluation noise across English and non-Hindi multilingual benchmarks. These results preserve model quality while recovering the efficiency motivation for vocabulary extension. Hindi fertility drops from $1.95$ to $1.25$, yielding $35.9\%$ fewer tokens and $24.5\%$ faster evaluation runtime from the reduced token count. These long-run results suggest that the asymmetric initialization remains stable at scale. We limit this run to roughly $50$B tokens due to compute constraints; additional CPT may further improve scores.

\section{Conclusion and Future Work}
This study systematically evaluates embedding initialization for LLM vocabulary extension. Our findings show that subword composition provides the largest improvement over vocabulary-averaging baselines, while input-side norm calibration and input-output decoupling further refine performance within the stronger subword-composition family. The best observed asymmetric strategy---applying uniform subword averaging with language-specific norm calibration to the input matrix ($\mathbf{E}_{\text{in}}$), and character-length weighting to the output head ($\mathbf{E}_{\text{out}}$)---substantially reduces the cold-start optimization cost relative to the default initialization baseline, achieving over a $6\times$ reduction in steps for comparable validation loss and a $7\times$ reduction for downstream benchmark performance.

Additionally, we highlight a methodological warning: initialization-time metrics can undergo ranking inversions during early CPT, making them unreliable proxies for later convergence. In contrast, a brief $50$-step CPT probe provides a stable and reliable signal for selecting the initialization strategy. Future work will investigate the generalizability of this asymmetric subword composition framework across other non-Latin writing systems, including Arabic, CJK scripts, and non-Indic language families.

\section*{Limitations}

This study focuses on Hindi vocabulary extension for Nemotron-3-Nano-30B-A3B, so the findings may not directly transfer to other languages, scripts, tokenizer families, or model architectures. The newly added Hindi tokens are sourced from the Nanda tokenizer, so we study embedding initialization rather than vocabulary construction. Finally, while the $50$-step CPT probe is reliable in our experiments, it should be validated across additional training mixtures and downstream benchmarks. Asymmetric methods achieve lower early loss values in our runs, but the differences among top configurations diminish as training progresses, implying that the asymmetry result should be interpreted as a practical design signal rather than a definitive separation among the strongest initializers.

\bibliography{main}

\clearpage
\onecolumn
\appendix

\section{Extended Convergence Curves}
\label{app:extended_curves}

\begin{figure}[!ht]
\centering
\includegraphics[width=\textwidth]{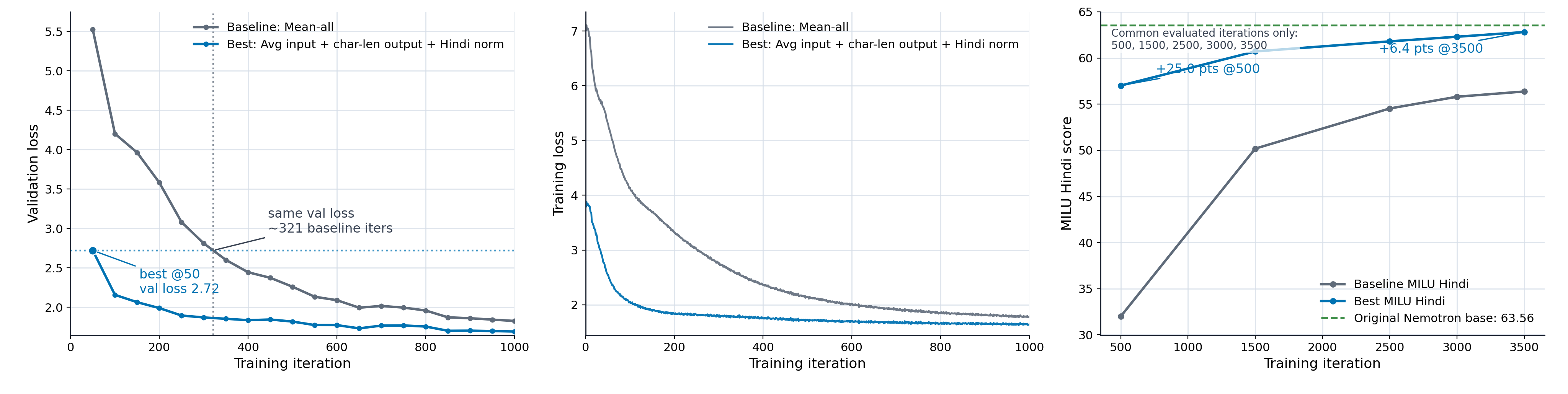}
\caption{Extended validation-loss, training-loss, and MILU-Hindi trajectories comparing the Mean-all baseline with the best observed initialization pipeline. The left panel shows that the best observed configuration reaches validation loss $2.72$ at step $50$, matching the Mean-all baseline after approximately $321$ iterations; the right panel shows the downstream MILU-Hindi comparison used for the $500$-step and $3{,}500$-step analysis.}
\label{fig:extended_loss_milu}
\end{figure}

\section{English and Multilingual Retention}
\label{app:retention_results}

Table~\ref{tab:retention_full} compares Original with Extended Best on English and non-Hindi multilingual benchmarks. All differences remain within roughly one standard error, indicating that Extended Best matches the original within evaluation noise. For code, raw counts change from $126/164$ to $123/164$ on HumanEval and from $191/257$ to $187/257$ on MBPP-Sanitized.

\begin{table}[!ht]
\centering
\small
\resizebox{\textwidth}{!}{
\begin{tabular}{llrrrr}
\noalign{\hrule height 1.3pt}
\textbf{Benchmark} & \textbf{Metric} & \textbf{Original} & \textbf{Extended Best} & \textbf{$\Delta$} & \textbf{$|\Delta|$/SE} \\
\noalign{\hrule height 1.3pt}
\multicolumn{6}{l}{\textbf{General knowledge \& reasoning}} \\ \hline
MMLU (5-shot) & \texttt{exact\_match} & 78.62 & 78.13 & $-0.49$ & 1.04 \\
MMLU-Pro (5-shot) & \texttt{exact\_match} & 64.53 & 64.88 & $+0.35$ & 0.59 \\
AGIEval-EN (CoT) & \texttt{exact\_match} & 67.75 & 67.84 & $+0.09$ & 0.07 \\
ARC-Challenge (25-shot) & \texttt{acc} & 91.81 & 91.64 & $-0.17$ & 0.15 \\
HellaSwag & \texttt{acc\_norm} & 85.57 & 85.12 & $-0.45$ & 0.90 \\
PIQA & \texttt{acc\_norm} & 84.33 & 83.68 & $-0.65$ & 0.54 \\
RACE & \texttt{acc} & 88.13 & 86.89 & $-1.24$ & 0.86 \\
WinoGrande (5-shot) & \texttt{acc} & 79.08 & 79.87 & $+0.79$ & 0.49 \\
OpenBookQA & \texttt{acc\_norm} & 46.40 & 46.80 & $+0.40$ & 0.13 \\ \hline
\multicolumn{6}{l}{\textbf{Code}} \\ \hline
HumanEval (greedy) & \texttt{pass@1} & 76.83 & 75.00 & $-1.83$ & 0.39 \\
MBPP-Sanitized (3-shot) & \texttt{pass@1} & 74.32 & 72.76 & $-1.56$ & 0.40 \\ \hline
\multicolumn{6}{l}{\textbf{Math}} \\ \hline
GSM8K (8-shot CoT) & \texttt{exact\_match} & 92.04 & 92.04 & $+0.00$ & 0.00 \\
MATH-500 (4-shot) & \texttt{pass@1[32]} & 77.46 & 77.04 & $-0.42$ & 0.17 \\ \hline
\multicolumn{6}{l}{\textbf{Multilingual (non-English, non-Hindi)}} \\ \hline
Global-MMLU-Lite (8-lang avg) & \texttt{acc} & 74.56 & 75.13 & $+0.57$ & 0.52 \\
MGSM (6-lang avg, CoT 8-shot) & \texttt{flexible-extr.} & 80.07 & 80.53 & $+0.47$ & 0.33 \\
\noalign{\hrule height 1.3pt}
\end{tabular}}
\caption{English and multilingual retention evaluation after the long-run CPT setting. All values are percentages; $\Delta$ is Extended Best minus Original.}
\label{tab:retention_full}
\end{table}

\begin{table}[!ht]
\centering
\small
\begin{tabular}{lrrr}
\noalign{\hrule height 1.3pt}
\textbf{Category} & \textbf{Original} & \textbf{Extended Best} & \textbf{$\Delta$} \\
\noalign{\hrule height 1.3pt}
General knowledge \& reasoning (9) & 76.25 & 76.09 & $-0.15$ \\
Code (2) & 75.58 & 73.88 & $-1.70$ \\
Math (2) & 84.75 & 84.54 & $-0.21$ \\
Multilingual (2) & 77.32 & 77.83 & $+0.52$ \\ \hline
Mean across all 15 benchmarks & --- & --- & $-0.28$ \\
Mean across 13 English-only benchmarks & --- & --- & $-0.40$ \\
\noalign{\hrule height 1.3pt}
\end{tabular}
\caption{Category-level retention summary for the benchmarks in Table~\ref{tab:retention_full}.}
\label{tab:retention_summary}
\end{table}

\end{document}